\documentclass{article} % For LaTeX2e
\usepackage{iclr2027_conference,times}

\usepackage{amsmath,amsfonts,bm}

\def\eqref#1{equation~\ref{#1}}
\def\1{\bm{1}}

\DeclareMathAlphabet{\mathsfit}{\encodingdefault}{\sfdefault}{m}{sl}
\SetMathAlphabet{\mathsfit}{bold}{\encodingdefault}{\sfdefault}{bx}{n}

\DeclareMathOperator*{\argmax}{arg\,max}

\usepackage{hyperref}
\usepackage{url}

\usepackage[T1]{fontenc}
\usepackage[utf8]{inputenc}
\usepackage{amsmath,amssymb,bm,graphicx,booktabs,multirow,array}
\usepackage{hyperref,url,placeins}

\newcommand{\matv}{Mature-V}

\newcommand{\daswave}{DAS-Wave}
\newtheorem{proposition}{Proposition}
\title{\centering
Shallow Queries, Mature Values:\\
Depth-Asynchronous Self-Speculation\\
for Looped Transformers
}

\author{%
\parbox[t]{\dimexpr\textwidth-2\tabcolsep\relax}{%
\centering\normalfont
\textbf{Guanghao Li}$^{1,2}$%
\thanks{Email: \texttt{ligh24@mails.tsinghua.edu.cn}.}\quad
\textbf{Zihan Su}$^{1}$\quad
\textbf{Hao Yu}$^{1}$\quad
\textbf{Jinyang Jiang}$^{3}$\quad
\textbf{Tao Ren}$^{4}$\\[4pt]
\textbf{Zehao Li}$^{4}$\quad
\textbf{Feng Lu}$^{5}$\quad
\textbf{Ming Tang}$^{2}$%
\thanks{Corresponding authors.}\quad
\textbf{Chun Yuan}$^{1}$\footnotemark[2]\\[7pt]
{\small
$^{1}$\,Tsinghua University\quad
$^{2}$\,Southern University of Science and Technology\\[2pt]
$^{3}$\,The University of Hong Kong\quad
$^{4}$\,Peking University\\[2pt]
$^{5}$\,Shenzhen University of Advanced Technology
}%
}%
}

\iclrfinalcopy % Uncomment for camera-ready version, but NOT for submission.
\begin{document}

\maketitle

\begin{abstract}
Looped Transformers reuse a shared block across recurrent depths, making
autoregressive decoding expensive because every generated token requires many
sequential recurrent passes. Self-speculative decoders reduce this cost by
drafting at an early depth and verifying at full depth, but typically bind
draft computation to prefix representations from the same recurrent depth.
We find that queries and keys approach their final-depth representations
earlier than values, and controlled prefix-channel interventions show that
mature values substantially improve shallow draft predictions.
Motivated by this asymmetry, we introduce
\textbf{Depth-Asynchronous Self-Speculation (DAS)}, which decouples the depth
of draft computation from the depth of verified-prefix representations it
reads. Its \textbf{Mature-V} primitive lets shallow queries retrieve full-depth
prefix values without additional recurrent computation.
We further develop \textbf{DAS-Wave}, which combines depth-asynchronous prefix
reads with carried parallel refinement, progressive block growth, and an
independent full-depth verifier.
Across four recurrent-model checkpoints and mathematics and code workloads,
DAS-Wave achieves \textbf{4.00--6.96$\times$} mean throughput speedup over
paired full-depth autoregressive decoding in the same inference stack.
These results identify prefix-information depth as an effective design axis
for recurrent self-speculation.
\end{abstract}
\section{Introduction}

Looped Transformers improve model performance by repeatedly applying a shared
block of layers across recurrent depth, scaling inference-time computation
without proportionally increasing model size
\citep{geiping2025scaling,li2025scout,cdb2026}.
This recurrent computation progressively refines hidden states, but also makes
autoregressive (AR) generation expensive: before producing each new token, the
model must execute many recurrent passes sequentially.
As recurrent depth grows, these serial passes become a major source of decoding
latency.

Speculative decoding offers a natural way to amortize this cost.
A cheaper drafter proposes future tokens, while the target model verifies
multiple proposals together
\citep{leviathan2023,chen2023spec}.
Self-speculative methods instead reuse a cheaper execution of the target itself
as the drafter
\citep{draftverify2024,layerskip2024}.
For looped Transformers, intermediate recurrent states provide a natural source
of such drafts
\citep{loopspec2026}.
Draft depth therefore forms a natural compute--accuracy tradeoff:
shallower exits are cheaper, but typically yield less reliable proposals.

We ask a complementary question:
\emph{what depth of prefix information should an early draft be allowed to
read?}
For verified prefix tokens, deeper target representations are already
available.
Nevertheless, an early draft typically reads cached prefix representations
from the same or a similarly shallow recurrent depth.
Thus, the computational depth of the draft is coupled to the representation
depth of the prefix it consumes.
This coupling is convenient, but it is not required by speculative
verification.

We find a systematic asymmetry in how attention representations mature across
recurrent depth.
Queries and keys approach their final-depth representations earlier than
values, suggesting that shallow computation can already form useful attention
queries while still retrieving relatively immature prefix content.
Controlled interventions confirm that this distinction matters for prediction:
replacing only the prefix values available to a shallow draft with their
full-depth counterparts substantially improves agreement with the full-depth
target, without increasing the draft's recurrent depth or requiring additional
draft computation passes
(Figure~\ref{fig:motivation}).

Motivated by this asymmetry, we introduce
\textbf{Depth-Asynchronous Self-Speculation (DAS)},
which decouples the recurrent depth at which a draft is computed from the depth
of verified-prefix representations it reads.
We instantiate this principle with
\textbf{\matv{}},
a training-free primitive that lets shallow draft queries retrieve mature
values already cached from deeper recurrent states of the verified prefix.
\matv{} improves the information available to an early draft without making
the draft itself deeper, and can be applied directly to conventional sequential
self-speculation without changing its verification procedure.

\begin{figure}[t]
\centering
\includegraphics[width=\linewidth]{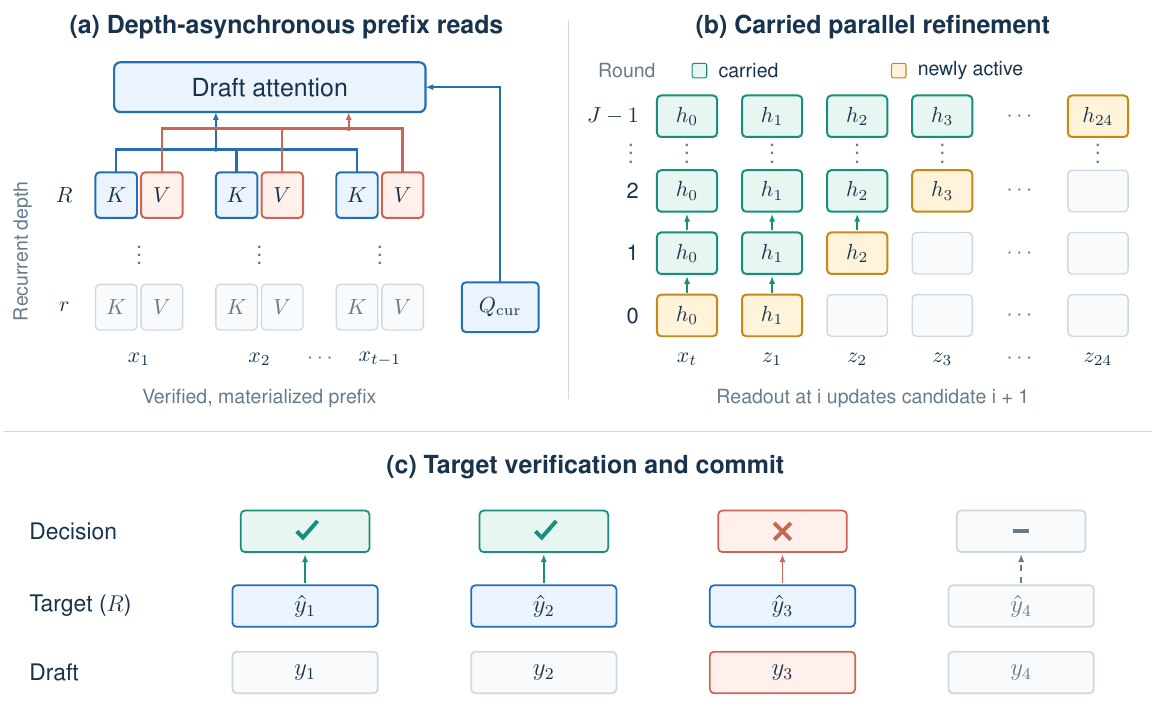}
\caption{
\textbf{DAS-Wave overview.}
(a) Queries from the current draft state attend to full-depth K/V from the
verified, materialized prefix at the same core layer.
(b) Active positions are refined in parallel while latent states are carried
across rounds and the active window grows. Gold marks newly activated states.
The anchor remains fixed, and each readout updates the next candidate position.
(c) An independent full-depth target verifies the candidates without using
carried draft states. In the illustrated example, $y_1$ and $y_2$ are accepted;
$y_3$ is the first mismatch, so the target prediction
$\hat y_3=\argmax_v p_3(v)$ is emitted and the remaining draft suffix is
discarded.
}
\label{fig:method}
\end{figure}

Better draft predictions alone, however, do not guarantee lower end-to-end
latency: their benefit must outweigh the cost of constructing a speculative
block.
To translate depth-asynchronous prefix reuse into acceleration, we further
develop
\textbf{\daswave{}},
an end-to-end decoder built on recurrent parallel refinement
\citep{psampler2025}.
\daswave{} refines multiple candidate positions in parallel, carries recurrent
states across refinement rounds, and progressively expands the speculative
block.
A separate full-depth target pass verifies the proposals and materializes the
deep prefix representations used by subsequent speculation.
Thus, \daswave{} changes proposal construction while leaving the target model
and standard speculative verification rule unchanged
(Figure~\ref{fig:method}).

Across four recurrent-model checkpoints and mathematics and code workloads,
\daswave{} achieves \textbf{4.00--6.96$\times$} mean throughput speedup over
paired full-depth autoregressive decoding.
Our primary contributions are:
\begin{itemize}
    \item \textbf{Depth-asynchronous prefix reuse.}
    We show that recurrent representations need not mature uniformly across
    channels, motivating the decoupling of draft depth from verified-prefix
    representation depth.

    \item \textbf{DAS, Mature-V, and DAS-Wave.}
    We introduce DAS and its training-free \matv{} primitive, and integrate
    depth-asynchronous prefix reads with carried parallel refinement and
    independent full-depth verification.

    \item \textbf{Mechanism and system evaluation.}
    Controlled channel and carry interventions explain the full decoder's
    prefix policy, while evaluations across four recurrent checkpoints
    establish \textbf{4.00--6.96$\times$} throughput speedup and characterize
    its computation tradeoffs.
\end{itemize}

\section{Background and related work}
\label{sec:background}

\paragraph{Looped and recurrent-depth Transformers.}
Transformers with computation shared across depth range from Universal
Transformers and early looped architectures to recent recurrent-depth language
models
\citep{dehghani2019universal,giannou2023looped,li2026efficient,
bae2025relaxed,mcleish2025retrofit}.
Rather than stacking distinct parameters at every depth, these models repeatedly
apply a shared computation, allowing effective depth---and hence inference-time
compute---to grow without proportionally increasing parameter count.
We write a looped language model as
\begin{equation}
e=\mathcal P(x),\qquad
h^{(r+1)}=\mathcal F(h^{(r)},e),\qquad
p_r=\operatorname{softmax}\!\left(\mathcal H(h^{(r)})\right),
\label{eq:recurrence}
\end{equation}
where $\mathcal P$ is the prelude, $\mathcal F$ the shared recurrent core,
$\mathcal H$ the coda, $r$ recurrent depth, and $R$ the full target depth.
At core layer $\ell$, a verified prefix $C$ has depth-indexed cached
representations $K_C^{r,\ell}$ and $V_C^{r,\ell}$.
Conventional depth-aligned drafting reads prefix K/V from the same recurrent
depth as the current draft.
DAS relaxes this recurrence-depth alignment while preserving the corresponding
core-layer index $\ell$.

\paragraph{Speculative and self-speculative decoding.}
Speculative decoding accelerates autoregressive generation by using a cheaper
proposal mechanism and evaluating multiple future tokens with the target in
parallel
\citep{leviathan2023,chen2023spec}.
Alternative proposal mechanisms exploit structure already available during
decoding: Lookahead Decoding constructs candidates from parallel Jacobi
trajectories \citep{lookahead2024}, prompt lookup reuses repeated context
substrings \citep{pld2023}, and Token Recycling and SAM-Decoding reuse
previously computed predictions or suffix structure
\citep{tokenrecycling2024,samd2025}.
Self-speculative methods instead obtain cheaper proposals from the target
itself, for example through layer skipping or early exits
\citep{draftverify2024,layerskip2024}.
Looped Transformers naturally expose intermediate recurrent predictions
\citep{geiping2025scaling}, and LoopSpec exploits them for training-free
self-speculative decoding with full-depth target verification
\citep{loopspec2026}.
Existing approaches primarily differ in how proposals are constructed or
refined.
DAS instead studies a complementary axis:
\emph{which recurrent-depth representation of the already verified prefix is
read by the draft}.

\paragraph{Parallel proposal generation for recurrent-depth models.}
SPEED speculatively executes future tokens in parallel using predictions from
early hidden states in parameter-sharing decoders
\citep{speed2023}.
More closely related to \daswave{}, Parallel Samplers exploit recurrent-depth
structure to refine multiple candidate positions concurrently while carrying
latent states across refinement rounds
\citep{psampler2025}.
We adopt this carried refinement and progressive window growth as the proposal
engine in \daswave{}.
Parallel Samplers primarily determine how candidate states are refined,
whereas DAS changes which recurrent-depth representations of the verified
prefix those candidates read.
\daswave{} combines these depth-asynchronous prefix reads with carried parallel
refinement and an independent full-depth target verifier.
Thus, carry and window growth are inherited from prior parallel sampling,
while depth-asynchronous prefix reuse and its interaction with verified
parallel drafting are the focus of this work.

\section{Depth-Asynchronous Self-Speculation}
\label{sec:method}

DAS decouples the recurrent depth of draft computation from the depth of the
verified-prefix representations it reads.
This separation allows a shallow draft to reuse information that has already
been refined by the full-depth target without increasing the draft's own
recurrent computation.
We first motivate this design through channel-wise depth analysis, then
formalize depth-asynchronous K/V reads and instantiate them in the parallel
\daswave{} decoder.

\subsection{Design motivation: shallow queries, mature values}
\label{sec:motivation}

\begin{figure}[t]
\centering
\includegraphics[width=\linewidth]{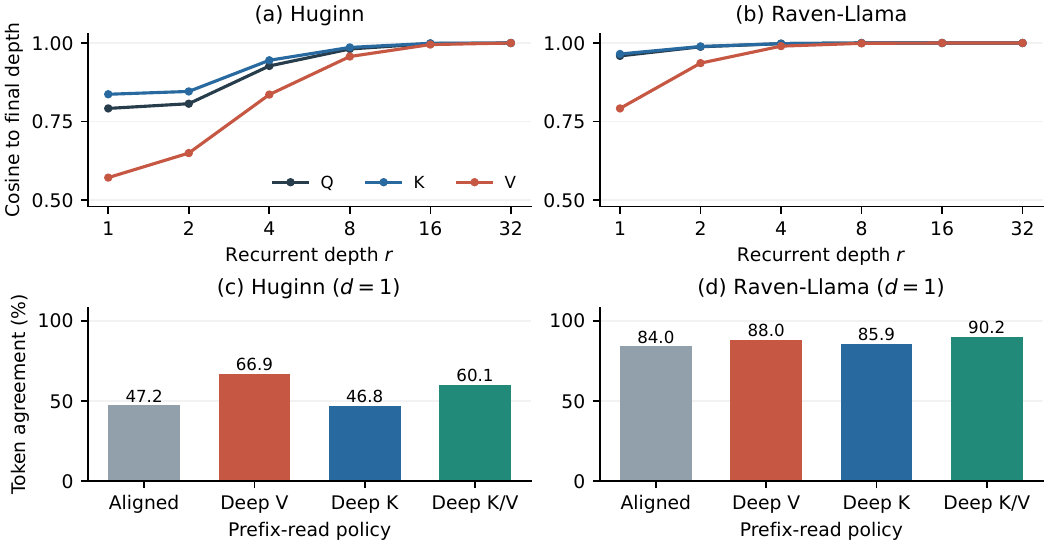}
\caption{
\textbf{Shallow queries, mature values.}
Top: Q/K/V cosine similarity to the final recurrence ($R=32$), averaged over
core layers.
Queries and keys approach their final-depth representations earlier than
values.
Bottom: draft--target token agreement under separate prefix-channel
replacements.
Deep V uses full-depth prefix V while keeping K aligned with the draft; in
sequential drafting, this corresponds to \matv{}.
Each model uses 12 GSM8K prompts and 756 evaluated next-token prediction
positions.
}
\label{fig:motivation}
\end{figure}

Figure~\ref{fig:motivation} examines how attention representations evolve
across recurrent depth.
At the first recurrence, mean cosine similarity to the final-depth
representation is $0.79/0.84/0.57$ for Q/K/V on Huginn and
$0.96/0.97/0.79$ on Raven-Llama.
Queries and keys therefore approach their final-depth representations earlier
than values, suggesting that shallow computation may already provide useful
routing signals while still retrieving relatively immature prefix content.

We test the predictive consequence by changing only the recurrent depth of the
verified prefix's cached attention channels.
At draft depth $r=1$, replacing prefix V with its full-depth counterpart raises
Huginn's draft--target agreement from $47.22\%$ to $66.93\%$, whereas
replacing K alone gives $46.83\%$.
On Raven-Llama, V-only replacement raises agreement from $83.99\%$ to
$87.96\%$.
Thus, mature prefix values can substantially improve shallow predictions
without increasing the draft's own recurrent depth.
This observation motivates \matv{} as the minimal depth-asynchronous
intervention: enrich the content retrieved by a shallow draft while leaving
its query computation shallow.
Complete channel and depth sweeps appear in
Appendix~\ref{app:channels}.

\subsection{Depth-asynchronous prefix reads}
\label{sec:prefix-read}

For an attention call at core layer $\ell$, let $C$ denote the materialized
verified prefix and $U$ the active draft-side positions.
DAS makes the verified-prefix read depth explicit through two choices:
$\kappa$ for keys and $\nu$ for values.
The attention weights are
\begin{equation}
[\Pi_C,\Pi_U]
=
\operatorname{softmax}\!\left(
Q^{\mathrm{cur},\ell}
[K_C^{\kappa,\ell};K_U^{\mathrm{cur},\ell}]^\top
/\sqrt{d_h}+M
\right),
\label{eq:policy-attn}
\end{equation}
and the corresponding attention output is
\begin{equation}
O^\ell
=
\Pi_CV_C^{\nu,\ell}
+
\Pi_UV_U^{\mathrm{cur},\ell}.
\label{eq:policy-output}
\end{equation}
Here, $M$ is the causal mask and $[\,;\,]$ concatenates positions.
Verified-prefix and active draft-side positions participate in the same
attention normalization.
Only the verified-prefix read depths are changed by DAS; draft-side K/V are
always produced by the current draft computation.

For a sequential draft at recurrent depth $r$, the verified-prefix read
policies are
\begin{equation}
\begin{array}{c@{\qquad}c}
\text{Aligned}:~(\kappa,\nu)=(r,r),
&
\text{\matv{} / Deep V}:~(\kappa,\nu)=(r,R),
\\[2pt]
\text{Deep K}:~(\kappa,\nu)=(R,r),
&
\text{Deep K/V}:~(\kappa,\nu)=(R,R).
\end{array}
\label{eq:read-policies}
\end{equation}
\matv{} keeps prefix K aligned with the shallow draft while reading prefix V
from full depth.
Throughout the channel ablations, we refer to this V-only assignment as
\textbf{Deep V}; in sequential drafting, Deep V and \matv{} therefore denote
the same read-depth assignment, with \matv{} naming the corresponding
depth-asynchronous primitive.

Equation~\ref{eq:read-policies} defines a $2\times2$ read-depth design space
over shallow versus full-depth prefix K/V.
The V-only intervention isolates the strongest single-channel effect in
sequential drafting.
Under carried parallel refinement, however, mature K and V become strongly
complementary (Section~\ref{sec:components}).
Because cross-depth reads are restricted to verified prefix tokens, their
full-depth representations have already been materialized by target
verification and require no additional recurrent passes.

\subsection{DAS-Wave: carried parallel refinement}
\label{sec:wave}

Depth-asynchronous reads improve individual proposals, but a conventional
self-speculative drafter still constructs a speculative block sequentially,
one proposed token after another.
\daswave{} shortens this proposal path by combining depth-asynchronous prefix
reads with carried parallel refinement, following the proposal structure of
recurrent Parallel Samplers \citep{psampler2025}.
Throughout \daswave{}, verified-prefix reads use the fixed Deep K/V policy
$(\kappa,\nu)=(R,R)$.

Let $x_t$ be the last committed token and fixed \emph{anchor}.
At refinement round $j$, the active width grows by one position per round:
\begin{equation}
W_j
=
\min(\gamma,W_{\mathrm{init}}+j),
\qquad
j=0,\ldots,J-1,
\label{eq:width}
\end{equation}
where $\gamma$ is the verifier width capacity, which also upper-bounds the
active refinement width, $W_{\mathrm{init}}$ is the initial active width
including the anchor, and $J$ is the number of refinement rounds.
We use $W_{\mathrm{init}}=2$ in the full decoder.

Let
$Z^{(j)}=[x_t,z_1^{(j)},\ldots,z_{W_j-1}^{(j)}]$
denote the active token guesses and
$H^{(j)}=[h_0^{(j)},\ldots,h_{W_j-1}^{(j)}]$
their latent states.
At initialization, provisional token guesses are seeded by repeating the
anchor token, so $W_{\mathrm{init}}=2$ starts from the anchor and one
provisional input.
This seed is only an initialization: the anchor's readout updates $z_1$,
while the readout from the last active position supplies the next candidate
before it enters the active window.
Newly activated latent states use the checkpoint's standard initialization.
Here $j$ indexes refinement rounds rather than ordinary target recurrent depth.

The materialized verified prefix $C$, preceding the anchor, provides the
full-depth cache
$\mathsf{Cache}_C^R=\{K_C^{R,\ell},V_C^{R,\ell}\}_{\ell}$.
Existing positions begin each round from their saved latent states.
All active positions then receive $k$ recurrent draft updates in parallel,
followed by a shifted token readout:
\begin{align}
H_{0:W_j-1}^{(j+1)}
&=
\mathcal F_{\mathrm{draft}}^{[k]}
\!\left(
H_{0:W_j-1}^{(j)},
\mathcal P(Z_{0:W_j-1}^{(j)});
\mathsf{Cache}_C^R
\right),
\nonumber\\
z_{i+1}^{(j+1)}
&=
\argmax_v[\mathcal H(h_i^{(j+1)})]_v,
\qquad
0\le i<W_j.
\label{eq:wave-update}
\end{align}
Here $\mathcal F_{\mathrm{draft}}^{[k]}$ denotes $k$ recurrent applications
under the DAS draft-attention policy.
Queries are produced from the current carried draft states, while
verified-prefix K/V are read from the mature cache.

The distinction between \emph{token guesses} and \emph{latent states} is
important.
A latent state remains attached to its input position and accumulates recurrent
computation across refinement rounds, even when the token guess used to
condition that position is revised.
Its readout updates the guess for the following candidate position.
Consequently, carry preserves intermediate computation instead of restarting
every active position whenever the candidate sequence changes.
Only newly activated positions begin from fresh states.
Temporary K/V associated with provisional draft positions are discarded
between rounds, whereas the carried latent states and verified full-depth
prefix cache are retained.

The carried state therefore should not be interpreted as an ordinary target
state at a particular recurrent depth $r$: its computation history spans
multiple refinement rounds and may be conditioned on successive candidate
guesses.
This altered proposal trajectory also changes how verified-prefix read depth
interacts with the draft.
Section~\ref{sec:components} shows that joint mature K/V becomes substantially
more effective under carry.

Together with the per-round recurrent-update count $k$, the schedule
parameters $J$, $\gamma$, and $W_{\mathrm{init}}$ determine the
proposal-computation budget.
After the final refinement round, let $w$ denote the number of draft proposals
passed to verification and write $y_i=z_i^{(J)}$ for $1\le i\le w$.
The anchor together with $y_1,\ldots,y_{w-1}$ forms the verifier input, while
the final readout supplies $y_w$.

\subsection{Independent full-depth verification}
\label{sec:verification}

The independent target consumes
$[x_t,y_1,\ldots,y_{w-1}]$ and executes all $R$ recurrent steps, producing
shifted next-token conditionals
\begin{equation}
p_i=p_R(\,\cdot\mid x_{\le t},y_{<i}),
\qquad 1\le i\le w.
\label{eq:verify-conditionals}
\end{equation}
Let $m\le w$ denote the number of proposals actually tested, and let
$\hat y_i=\argmax_v p_i(v)$ denote the corresponding target prediction.
Because $p_w$ predicts the token following $y_{w-1}$, it can verify $y_w$
even though $y_w$ is not itself part of the verifier input.

For greedy decoding, proposals are accepted in order while
$y_i=\hat y_i$.
At the first mismatch, $\hat y_i$ is emitted and the remaining speculative
suffix is discarded.
Figure~\ref{fig:method}(c) illustrates this case:
$y_1$ and $y_2$ are accepted, $y_3$ is the first mismatch, and
$\hat y_3$ is emitted in its place.
If all $m$ tested proposals agree, an additional target token is available
only when $m<w$; when $m=w$, every returned conditional has already been used
to verify a proposal.

The verifier uses the target model's ordinary initialization and full-depth
computation and receives no carried draft states.
Full-depth K/V corresponding to accepted, materialized inputs are retained as
the verified prefix, while cache entries associated with rejected candidates
are discarded.
A final emitted token that has been predicted but not yet processed as a
target input becomes the anchor of the next block and is materialized by the
subsequent verification pass.

This verification step closes the DAS-Wave loop: the full-depth target both
determines the committed output and supplies the mature prefix cache consumed
by the next draft.
Because DAS modifies only proposal construction and prefix reads while leaving
the target computation and verification rule unchanged, greedy target
preservation follows the standard speculative-decoding argument in exact
arithmetic \citep{leviathan2023}.

\section{Experiments}
\label{sec:experiments}

\subsection{Setup}

\paragraph{Models and tasks.}
We evaluate Huginn-0125 (3.5B) \citep{geiping2025scaling} and three
Raven checkpoints---Raven-Llama-3.2, Raven-OLMo-2, and Raven-TinyLlama
\citep{mcleish2025retrofit}---all with $R=32$ recurrent steps.
We use GSM8K \citep{cobbe2021}, MATH-500
\citep{hendrycks2021,lightman2023}, HumanEval \citep{chen2021code},
and MBPP \citep{austin2021}, with 40 prompts per model--task setting.
All main latency measurements use batch size 1 on a single NVIDIA L20X GPU.
Prompt templates, stopping rules, generation limits, and other protocol
details are provided in Appendix~\ref{app:protocol}.

\paragraph{Baselines and evaluation.}
We compare full-depth autoregressive decoding (AR) with Lookahead Decoding
\citep{lookahead2024}, prompt lookup decoding (PLD) \citep{pld2023},
Token Recycling \citep{tokenrecycling2024}, SAM-Decoding \citep{samd2025},
sequential self-speculation \citep{geiping2025scaling,loopspec2026},
Mature-Spec, and \daswave{}.
Mature-Spec retains sequential self-speculation while replacing aligned
prefix reads with the fixed Deep K/V policy.
Our full \daswave{} configuration uses $\gamma=32$, $J=24$, and
$W_{\mathrm{init}}=2$, with one-position growth per refinement round and
$k=2$ recurrent updates per round for Huginn and $k=1$ for the Raven
checkpoints.
All main decoding results use greedy decoding.
We report mean throughput speedup over the corresponding paired full-depth AR
baseline, including prefill, together with accepted length $A$, the mean
number of draft tokens accepted per verification.
Baseline-specific settings and configuration selection are reported in
Appendix~\ref{app:configs}, while metric definitions and confidence intervals
are detailed in Appendix~\ref{app:metrics}.

\begin{table}[t]
\centering
\fontsize{8.7}{10.4}\selectfont
\caption{
\textbf{End-to-end decoding performance.}
Speedup is mean throughput relative to paired full-depth AR; $A$ is the mean
number of accepted draft tokens per verification.
Bold marks the largest speedup and accepted length in each setting.
}
\label{tab:system}
\setlength{\tabcolsep}{2.4pt}

\begin{tabular*}{\linewidth}{@{\extracolsep{\fill}}ll*{4}{rr}@{}}
\toprule
 & & \multicolumn{2}{c}{GSM8K}
 & \multicolumn{2}{c}{MATH-500}
 & \multicolumn{2}{c}{HumanEval}
 & \multicolumn{2}{c}{MBPP} \\
\cmidrule(lr){3-4}
\cmidrule(lr){5-6}
\cmidrule(lr){7-8}
\cmidrule(l){9-10}
Model & Method
& Speedup & $A$
& Speedup & $A$
& Speedup & $A$
& Speedup & $A$ \\
\midrule

\multirow{7}{*}{Huginn}
& Lookahead
    & 1.83$\times$ & 1.1
    & 1.84$\times$ & 1.0
    & 1.50$\times$ & 0.6
    & 1.40$\times$ & 0.5 \\
& PLD
    & 1.54$\times$ & 0.7
    & 1.93$\times$ & 1.2
    & 1.61$\times$ & 0.8
    & 1.35$\times$ & 0.5 \\
& Token Recycling
    & 2.81$\times$ & 2.1
    & 2.81$\times$ & 2.1
    & 2.74$\times$ & 2.1
    & 2.56$\times$ & 1.8 \\
& SAM-Decoding
    & 1.63$\times$ & 0.9
    & 2.50$\times$ & 1.9
    & 2.01$\times$ & 1.3
    & 1.49$\times$ & 0.6 \\
& Self-speculative
    & 1.70$\times$ & 5.2
    & 1.82$\times$ & 5.7
    & 1.64$\times$ & 4.6
    & 1.79$\times$ & 5.0 \\
& Mature-Spec
    & 2.26$\times$ & 4.9
    & 2.41$\times$ & 5.4
    & 2.35$\times$ & 4.8
    & 2.31$\times$ & 4.6 \\
& \daswave{}
    & \textbf{4.12$\times$} & \textbf{15.6}
    & \textbf{4.64$\times$} & \textbf{17.3}
    & \textbf{4.00$\times$} & \textbf{14.9}
    & \textbf{4.20$\times$} & \textbf{15.5} \\
\midrule

\multirow{7}{*}{\shortstack[l]{Raven\\Llama}}
& Lookahead
    & 1.69$\times$ & 1.1
    & 1.56$\times$ & 0.8
    & 1.40$\times$ & 0.6
    & 1.46$\times$ & 0.7 \\
& PLD
    & 1.34$\times$ & 0.5
    & 2.14$\times$ & 1.4
    & 1.64$\times$ & 0.8
    & 2.10$\times$ & 1.4 \\
& Token Recycling
    & 2.59$\times$ & 2.0
    & 2.51$\times$ & 1.9
    & 2.42$\times$ & 1.8
    & 2.49$\times$ & 1.9 \\
& SAM-Decoding
    & 1.53$\times$ & 0.7
    & 2.34$\times$ & 1.7
    & 1.73$\times$ & 0.9
    & 2.43$\times$ & 1.7 \\
& Self-speculative
    & 3.12$\times$ & 11.4
    & 3.19$\times$ & 11.8
    & 3.07$\times$ & 10.8
    & 3.53$\times$ & 12.6 \\
& Mature-Spec
    & 3.50$\times$ & 13.1
    & 3.70$\times$ & 14.0
    & 3.44$\times$ & 13.1
    & 3.84$\times$ & 14.7 \\
& \daswave{}
    & \textbf{5.36$\times$} & \textbf{17.0}
    & \textbf{6.01$\times$} & \textbf{18.9}
    & \textbf{5.21$\times$} & \textbf{16.3}
    & \textbf{6.28$\times$} & \textbf{19.7} \\
\midrule

\multirow{7}{*}{\shortstack[l]{Raven\\OLMo}}
& Lookahead
    & 1.87$\times$ & 1.0
    & 1.70$\times$ & 0.8
    & 1.33$\times$ & 0.4
    & 1.84$\times$ & 1.0 \\
& PLD
    & 1.44$\times$ & 0.6
    & 2.19$\times$ & 1.4
    & 1.58$\times$ & 0.8
    & 2.09$\times$ & 1.3 \\
& Token Recycling
    & 2.80$\times$ & 2.0
    & 2.79$\times$ & 2.0
    & 2.40$\times$ & 1.6
    & 3.19$\times$ & 2.5 \\
& SAM-Decoding
    & 1.65$\times$ & 0.8
    & 1.72$\times$ & 0.9
    & 2.12$\times$ & 1.5
    & 1.94$\times$ & 1.3 \\
& Self-speculative
    & 3.31$\times$ & 12.2
    & 3.30$\times$ & 12.1
    & 2.82$\times$ & 9.6
    & 3.60$\times$ & 12.7 \\
& Mature-Spec
    & 3.63$\times$ & 13.5
    & 3.73$\times$ & 13.8
    & 3.26$\times$ & 12.1
    & 3.90$\times$ & 14.8 \\
& \daswave{}
    & \textbf{5.54$\times$} & \textbf{17.3}
    & \textbf{5.96$\times$} & \textbf{18.1}
    & \textbf{4.82$\times$} & \textbf{14.6}
    & \textbf{6.34$\times$} & \textbf{19.4} \\
\midrule

\multirow{7}{*}{\shortstack[l]{Raven\\TinyLlama}}
& Lookahead
    & 1.92$\times$ & 1.2
    & 1.47$\times$ & 0.7
    & 1.52$\times$ & 0.7
    & 1.62$\times$ & 0.9 \\
& PLD
    & 1.95$\times$ & 1.2
    & 1.47$\times$ & 0.6
    & 2.64$\times$ & 2.0
    & 2.18$\times$ & 1.4 \\
& Token Recycling
    & 2.78$\times$ & 2.3
    & 2.48$\times$ & 1.9
    & 2.45$\times$ & 1.9
    & 2.56$\times$ & 2.0 \\
& SAM-Decoding
    & 1.88$\times$ & 1.1
    & 1.75$\times$ & 1.0
    & 2.78$\times$ & 2.2
    & 2.45$\times$ & 1.8 \\
& Self-speculative
    & 3.44$\times$ & 12.1
    & 3.42$\times$ & 12.1
    & 3.26$\times$ & 10.8
    & 3.72$\times$ & 12.5 \\
& Mature-Spec
    & 3.78$\times$ & 13.6
    & 3.77$\times$ & 13.6
    & 3.63$\times$ & 12.9
    & 4.04$\times$ & 14.5 \\
& \daswave{}
    & \textbf{6.34$\times$} & \textbf{18.1}
    & \textbf{6.39$\times$} & \textbf{18.0}
    & \textbf{6.02$\times$} & \textbf{17.0}
    & \textbf{6.96$\times$} & \textbf{19.6} \\
\bottomrule
\end{tabular*}
\end{table}

\subsection{Decoding performance}

Table~\ref{tab:system} reports the end-to-end results.
Across four recurrent-model checkpoints and four mathematics and code
workloads, \daswave{} achieves $4.00$--$6.96\times$ throughput speedup over
paired full-depth AR and obtains the largest measured speedup in all sixteen
settings.
It also produces the longest accepted blocks in every setting.
On GSM8K, \daswave{} accepts $15.6$--$18.1$ draft tokens per verification,
compared with $5.2$--$12.2$ for the tuned sequential self-speculative
baselines.
Thus, the throughput gains are accompanied by a substantial increase in the
amount of useful draft work retained by each full-depth target pass. The improvement is consistent across Huginn and all three Raven checkpoints,
despite their different sequential self-speculation strengths, indicating that
the gain is not tied to a single recurrent architecture or workload.

The progression from self-speculation to Mature-Spec and then to \daswave{}
shows how mature prefix reuse translates into system-level gains.
At their selected configurations, Mature-Spec improves over conventional
sequential self-speculation in all sixteen settings, showing that mature
prefix information is already beneficial within sequential self-speculation.
Mature-Spec changes the prefix-read policy while retaining sequential proposal
construction; \daswave{} further combines the same fixed Deep K/V policy with
carried parallel refinement and obtains substantially larger speedups.
Section~\ref{sec:components} isolates these two effects and explains why joint
Deep K/V is preferred under carried refinement.

\subsection{Prefix reads and latent carry}
\label{sec:components}

\textbf{Mature reads help in sequential drafting.}
Table~\ref{tab:components} separates prefix-read policy from latent carry under
controlled drafting settings.
Within sequential drafting, replacing aligned reads with Deep K/V improves
both accepted length and speed in all four model--task settings shown.
For example, on Huginn/GSM8K, $A$ increases from $2.80$ to $4.92$ and
speedup from $1.54\times$ to $2.26\times$.
Thus, mature prefix information is already useful before introducing parallel
proposal refinement.

\textbf{Mature reads and latent carry are complementary.}
The parallel controls show that the two mechanisms interact strongly.
Holding the refinement schedule fixed, either intervention alone changes
performance only modestly, whereas combining carry with Deep K/V yields much
larger gains than either alone.
On Huginn/GSM8K, the aligned/no-carry configuration runs at $0.65\times$;
Deep K/V without carry reaches $0.93\times$, and carry with aligned reads
reaches $0.71\times$, while their combination reaches $4.11\times$ with
$A=15.60$.
The same interaction appears on Raven-Llama/GSM8K: the two single
interventions reach $2.64\times$ and $2.66\times$, compared with
$5.40\times$ when Deep K/V and carry are combined.
MATH-500 exhibits the same qualitative pattern.
The carried Deep K/V configuration is the full \daswave{} decoder.

\textbf{Carry favors joint mature K/V.}
The carried channel controls also explain why \daswave{} uses joint Deep K/V
rather than Deep V alone.
With carry fixed, Deep V improves over aligned reads in all four settings,
consistent with the V-centric mechanism identified in
Figure~\ref{fig:motivation}.
Joint Deep K/V is substantially stronger still: on Huginn/GSM8K, Aligned,
Deep V, and Deep K/V give $0.71\times$, $1.07\times$, and $4.11\times$
speedup, with $A=1.61$, $2.99$, and $15.60$, respectively.
Thus, \matv{} isolates the benefit of mature values in shallow sequential
drafting, whereas the altered proposal trajectory under carry favors joint
mature K/V as the full-system policy.
Complete carry/K/V factorial analyses and independently retuned controls are
reported in Appendix~\ref{app:components}.

\begin{table}[t]
\centering
\fontsize{8.7}{10.4}\selectfont
\caption{
\textbf{Fixed-setting component comparisons.}
Sequential rows use draft depth 4.
All parallel rows share the same refinement schedule; no-carry rows
reinitialize latent states between rounds, while carried rows retain them.
Parallel + carry with Deep K/V uses the full \daswave{} configuration.
Component estimates are measured in their own paired timing sessions.
}
\label{tab:components}
\setlength{\tabcolsep}{3pt}

\begin{tabular*}{\linewidth}{@{\extracolsep{\fill}}lll*{2}{cc}@{}}
\toprule
 & & & \multicolumn{2}{c}{GSM8K}
 & \multicolumn{2}{c}{MATH-500} \\
\cmidrule(lr){4-5}
\cmidrule(l){6-7}
Model & Procedure & Prefix
& Speedup & $A$
& Speedup & $A$ \\
\midrule

\multirow{7}{*}{Huginn}
& Sequential
& Aligned
& 1.54$\times$ & 2.80
& 1.85$\times$ & 3.69 \\

& Sequential
& Deep K/V
& 2.26$\times$ & 4.92
& 2.42$\times$ & 5.35 \\

& Parallel, no carry
& Aligned
& 0.65$\times$ & 1.38
& 1.16$\times$ & 3.37 \\

& Parallel, no carry
& Deep K/V
& 0.93$\times$ & 2.42
& 1.06$\times$ & 2.95 \\

& Parallel + carry
& Aligned
& 0.71$\times$ & 1.61
& 1.20$\times$ & 3.51 \\

& Parallel + carry
& Deep V
& 1.07$\times$ & 2.99
& 1.66$\times$ & 5.26 \\

& Parallel + carry
& Deep K/V
& 4.11$\times$ & 15.60
& 4.63$\times$ & 17.30 \\
\midrule

\multirow{7}{*}{\shortstack[c]{Raven\\Llama}}
& Sequential
& Aligned
& 3.12$\times$ & 11.45
& 3.18$\times$ & 11.84 \\

& Sequential
& Deep K/V
& 3.50$\times$ & 13.13
& 3.67$\times$ & 14.01 \\

& Parallel, no carry
& Aligned
& 2.13$\times$ & 5.78
& 2.97$\times$ & 8.59 \\

& Parallel, no carry
& Deep K/V
& 2.64$\times$ & 7.50
& 3.34$\times$ & 9.84 \\

& Parallel + carry
& Aligned
& 2.66$\times$ & 7.60
& 3.16$\times$ & 9.28 \\

& Parallel + carry
& Deep V
& 3.52$\times$ & 10.45
& 4.52$\times$ & 13.84 \\

& Parallel + carry
& Deep K/V
& 5.40$\times$ & 16.97
& 5.99$\times$ & 18.88 \\
\bottomrule
\end{tabular*}
\end{table}

\subsection{Refinement cost--acceptance tradeoff}
\label{sec:sensitivity}

\begin{figure}[t]
\centering
\includegraphics[width=\linewidth]{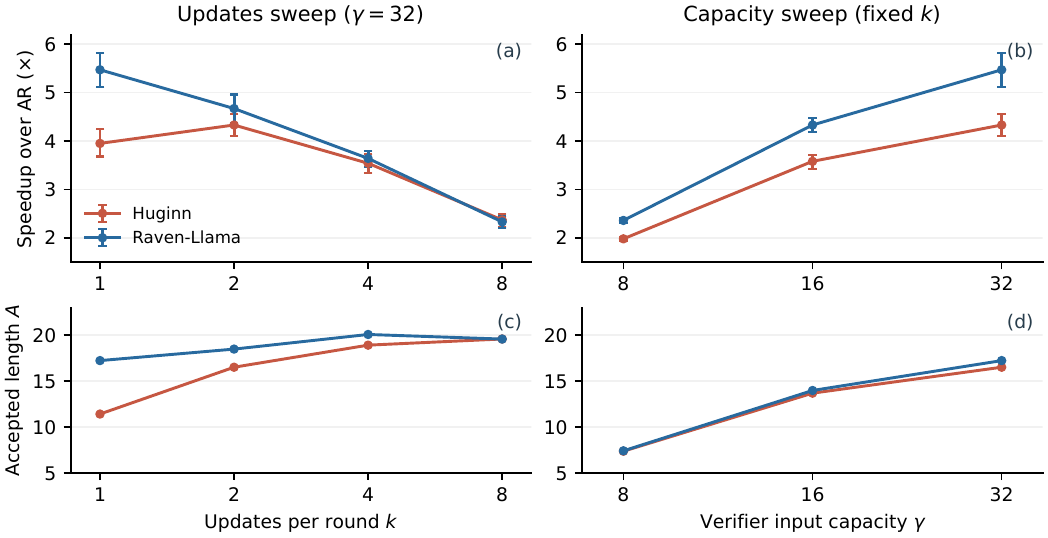}
\caption{
\textbf{Refinement cost--acceptance tradeoff.}
GSM8K diagnostic runs with $J=24$.
Top: mean throughput speedup; bottom: accepted draft length $A$.
Left varies recurrent updates per round; right varies verifier capacity at the
selected update count.
}
\label{fig:sensitivity}
\end{figure}

Figure~\ref{fig:sensitivity} examines how additional proposal computation
trades off against accepted work.
For Huginn, increasing recurrent updates per refinement round from $2$ to $8$
raises $A$ from $16.49$ to $19.56$, but reduces speedup from
$4.33\times$ to $2.37\times$.
Thus, better-refined proposals do not necessarily yield faster decoding:
their additional accepted tokens must compensate for the recurrent computation
spent producing them.
Raven-Llama reaches its highest speed at one update per round, consistent with
its stronger shallow predictions.

The capacity sweep shows the complementary effect.
At the selected update count, increasing capacity from $8$ to $32$ raises
speedup from $1.98\times$ to $4.33\times$ on Huginn and from
$2.36\times$ to $5.47\times$ on Raven-Llama.
A wider verifier can accommodate more candidate tokens, but additional
capacity is useful only when the refinement schedule activates and sufficiently
refines those positions.
Together, these results show that \daswave{} performance is governed by the
balance between accepted work and proposal cost, rather than accepted length
or nominal capacity alone.
Complete numerical sweeps and additional sensitivity to the initial active
width are reported in Appendix~\ref{app:budgets}.

\subsection{Applicability of DAS and DAS-Wave}
\label{sec:scope}

DAS naturally extends recurrent-depth self-speculation whenever intermediate
predictions and depth-indexed K/V representations of the verified prefix
are available.
Because it changes only the representations read by the draft, the same
principle can be used within conventional sequential self-speculation
without modifying target verification.
Depth-asynchronous reuse is especially useful when inexpensive intermediate
drafts coexist with prefix representations that continue to refine across
recurrent depth.
DAS-Wave further combines these depth-asynchronous reads with
input-conditioned recurrent refinement.
Carried latent states can incorporate revised token guesses across rounds,
allowing proposal computation to accumulate as the candidate sequence
is refined.

\section{Conclusion}
\label{main:end}

We introduced Depth-Asynchronous Self-Speculation (DAS), which decouples the
depth of draft computation from the depth of the verified-prefix
representations it reads.
Mature-V isolates a simple mechanism: shallow drafts can benefit from
full-depth values already materialized for the verified prefix.
DAS-Wave combines depth-asynchronous prefix reuse with carried parallel
refinement.
Across four recurrent-model checkpoints and mathematics and code workloads,
DAS-Wave achieves $4.00$--$6.96\times$ mean throughput speedup over paired
full-depth autoregressive decoding.
More broadly, our results identify prefix-information depth as a distinct
design axis for recurrent self-speculation:
draft computation need not be restricted to prefix representations of the
same computational age.
This separation suggests that proposal cost and prefix information quality can
be optimized independently, opening a broader design space for combining cheap
draft computation with mature information already produced by target
verification.

\clearpage
\appendix
\section{Experimental protocol}
\label{app:protocol}

This appendix provides the protocol details needed to reproduce the main
experiments, followed by algorithmic details, mechanism controls, refinement-budget studies, and performance diagnostics.
Each diagnostic uses its stated prompt set and paired AR reference; estimates
from separate timing sessions are not pooled.

\subsection{Models and tasks}

All main experiments use $R=32$ recurrent steps and the checkpoint's target
initialization convention.
A core layer denotes one layer inside the repeatedly applied shared block; its
index is distinct from recurrent depth.
Table~\ref{tab:models} summarizes the evaluated checkpoints.

\begin{table}[htbp]
\centering
\small
\caption{Evaluated recurrent-depth models.}
\label{tab:models}
\begin{tabular}{lrrrr}
\toprule
Checkpoint & Prelude & Shared core & Coda & Recurrences \\
\midrule
Huginn-0125 & 2 & 4 & 2 & 32 \\
Raven-Llama-3.2 & 4 & 6 & 4 & 32 \\
Raven-OLMo-2 & 4 & 6 & 4 & 32 \\
Raven-TinyLlama & 4 & 8 & 4 & 32 \\
\bottomrule
\end{tabular}
\end{table}

The main evaluation uses 40 fixed prompts per model--task setting, shared across
methods and disjoint from the prompts used for configuration selection.
GSM8K uses eight worked examples, MATH-500 uses four fixed Minerva-style
demonstrations \citep{lewkowycz2022}, and HumanEval and MBPP use
code-completion prompts.
Table~\ref{tab:tasks} summarizes the generation caps.
Generation stops at the earliest applicable EOS, task-specific stop condition,
or maximum generation length.

\begin{table}[htbp]
\centering
\small
\caption{Prompt conventions and generation caps.}
\label{tab:tasks}
\begin{tabular}{lrl}
\toprule
Task & Cap & Prompt convention \\
\midrule
GSM8K & 512 & Eight-shot chain of thought \\
MATH-500 & 1024 & Four-shot worked solutions \\
HumanEval & 512 & Code completion \\
MBPP & 512 & Code completion \\
\bottomrule
\end{tabular}
\end{table}

All main latency measurements use batch size 1 and bf16 on a single
NVIDIA L20X GPU.

\subsection{Draft configurations and baselines}
\label{app:configs}

The sequential self-speculative drafter is controlled by recurrent draft depth
$d$ and speculative block length $b$.
The development grid considers
$d\in\{4,8\}$ and $b\in\{8,16,32\}$, after which the selected settings are
frozen for evaluation.
The component experiments instead fix $d=4$ so that prefix-read policies are
compared at the same draft depth.

Table~\ref{tab:draft-config} summarizes the neural drafting configurations
used in the main experiments.
Mature-Spec uses the fixed Deep K/V policy, whereas conventional
self-speculation uses aligned prefix reads.
DAS-Wave uses the one-position growth schedule in
Equation~\ref{eq:width}.

\begin{table}[htbp]
\centering
\small
\caption{Frozen neural-draft configurations.}
\label{tab:draft-config}
\begin{tabular}{llll}
\toprule
Method & Huginn & Raven family & Prefix reads \\
\midrule
Self-speculative & $d=8,b=8$ & $d=4,b=16$ & Aligned \\
Mature-Spec & $d=4,b=8$ & $d=4,b=16$ & Deep K/V \\
DAS-Wave & $\gamma=32,k=2$ & $\gamma=32,k=1$ & Deep K/V \\
\bottomrule
\end{tabular}
\end{table}

All DAS-Wave rows use $J=24$ and $W_{\mathrm{init}}=2$.
The fixed-setting parallel controls use the same schedule and vary only latent
carry and the verified-prefix read policy.

The external baselines use their corresponding proposal mechanisms within the
same reference stack as their paired AR baseline.
Lookahead uses parallel Jacobi refinement and an $n$-gram candidate pool
\citep{lookahead2024}; PLD performs prompt lookup \citep{pld2023};
Token Recycling reuses previously predicted successor structure
\citep{tokenrecycling2024}; and SAM-Decoding uses dynamic suffix retrieval
\citep{samd2025}.
Construction, lookup, and update costs are included in timing.

\subsection{Timing, acceptance, and uncertainty}
\label{app:metrics}

For elapsed time $t_{p,a}$ and retained output-token count $n_{p,a}$ of method
$a$ on prompt $p$, the reported speedup is
\begin{equation}
S_a
=
\frac{1}{N}
\sum_{p=1}^{N}
\frac{n_{p,a}/t_{p,a}}
     {n_{p,\mathrm{AR}}/t_{p,\mathrm{AR}}}.
\label{eq:speed}
\end{equation}
CUDA-event timing includes both prefill and decoding.
Each accelerated arm is paired with a full-depth AR reference from the same
timing session and inference stack.

For verification event $b$, let $a_b$ denote the number of accepted draft
tokens retained after verification.
For prompt $p$ with $B_p>0$ verification events,
\begin{equation}
A_p
=
\frac{\sum_b a_b}{B_p},
\qquad
A=\operatorname{mean}_p A_p .
\label{eq:acceptance}
\end{equation}
Thus, $A$ is the mean accepted draft length per verification rather than an
acceptance probability.
Prefill contributes to elapsed time but not to verification-event counts.

Unless otherwise stated, 95\% confidence intervals are obtained from
10,000 bootstrap resamples over paired prompts.
For methods whose candidate store persists across requests, we bootstrap
contiguous blocks of five requests to preserve local stream dependence.
Direct method comparisons always use paired measurements from the same prompts
and timing session.

\section{Algorithm and target preservation}
\label{app:algorithm}

\subsection{State and cache lifetimes}

DAS-Wave maintains three distinct objects:
candidate token guesses, carried draft latent states, and the target's
depth-indexed prefix K/V.
Only target-materialized verified positions supply mature prefix reads.
Active draft-side positions use K/V produced by the current draft computation,
and these provisional K/V are discarded between refinement rounds.

The final emitted token of a block can be committed before it has been
processed as a full-depth target input.
This token becomes the fixed anchor of the next block.
It conditions drafting and verification but does not supply mature K/V until
the subsequent target pass materializes it.

\subsection{One DAS-Wave block}

Given a materialized prefix cache and anchor $x_t$, one block proceeds as
follows:
\begin{enumerate}
    \item \textbf{Initialize.}
    Initialize provisional token guesses by repeating the anchor and initialize
    newly active latent states using the checkpoint's convention.

    \item \textbf{Refine.}
    For $j=0,\ldots,J-1$, activate the first $W_j$ draft-side positions,
    perform $k$ recurrent updates using Deep K/V verified-prefix reads, and
    shift each position's prediction to the following candidate according to
    Equation~\ref{eq:wave-update}.

    \item \textbf{Carry.}
    Retain the updated latent states for the next refinement round while
    discarding provisional draft K/V.

    \item \textbf{Verify.}
    Run the independent full-depth target on
    $[x_t,y_1,\ldots,y_{w-1}]$.

    \item \textbf{Commit.}
    Accept proposals in order until the first mismatch, emit the target
    prediction at that position, and discard the remaining speculative suffix.
    Apply the same stopping rule and output cap as AR.
\end{enumerate}

With the default $W_{\mathrm{init}}=2$, the first round processes the anchor
and one provisional input.
The readout from the last active position supplies the next candidate before
that position becomes active in the following round.
The verifier capacity $\gamma$, active refinement width $W_j$, number of
tested proposals, and number of emitted tokens are therefore distinct
quantities.

\subsection{Greedy target preservation}
\label{app:theory}

\begin{proposition}[Greedy target preservation]
Fix a deterministic full-depth target, its initialization, tie-breaking rule,
and stopping rule.
If every verification pass evaluates the same target conditionals as
autoregressive decoding and retains the corresponding accepted-prefix target
cache, speculative verification emits the same greedy sequence as target AR
in exact arithmetic, regardless of the proposal procedure.
\end{proposition}

\paragraph{Proof.}
Assume that the committed prefix agrees with target AR through position $t$.
The first verifier conditional is therefore the target conditional on the same
prefix.
Any accepted proposal equals the argmax of that conditional and hence equals
the next AR token.
Repeating the argument covers every accepted proposal.
At the first disagreement, the verifier emits the target argmax, which again
equals the next AR token.
If an unused all-accepted conditional is available, its argmax is the next AR
token by the same argument.
Retaining only valid target cache entries and applying the same stopping rule
preserve the induction hypothesis for the next block.
Induction from the prompt proves the claim
\citep{leviathan2023}.

The proposition does not require the carried draft trajectory to coincide
with an ordinary target recurrence. As in standard speculative decoding,
finite-precision execution may introduce numerical differences
\citep{chen2023spec}.

\FloatBarrier
\section{Mechanism and component analyses}
\label{app:components}

\subsection{Depth-dependent channel replacement}
\label{app:channels}

The channel probe follows the full-depth target trajectory and changes only
the verified-prefix reads available to a shallow draft.
Each model contributes 756 next-token prediction positions from 12 GSM8K
prompts.
For each tested draft depth, we evaluate aligned, Deep V, Deep K, and
Deep K/V reads.
Table~\ref{tab:channels} reports the complete results.

\begin{table}[htbp]
\centering
\small
\caption{Draft--target token agreement under depth-dependent prefix-channel
substitution.}
\label{tab:channels}
\setlength{\tabcolsep}{6pt}
\begin{tabular}{@{}llrrrr@{}}
\toprule
Model & Prefix policy & $d=1$ & $d=2$ & $d=4$ & $d=8$ \\
\midrule
\multirow{4}{*}{Huginn}
& Aligned  & 0.4722 & 0.6362 & 0.7937 & 0.9127 \\
& Deep V   & 0.6693 & 0.7659 & 0.8955 & 0.9484 \\
& Deep K   & 0.4683 & 0.6045 & 0.8016 & 0.8968 \\
& Deep K/V & 0.6005 & 0.7937 & 0.8995 & 0.9669 \\
\midrule
\multirow{4}{*}{\shortstack[l]{Raven\\Llama}}
& Aligned  & 0.8399 & 0.9034 & 0.9590 & 0.9841 \\
& Deep V   & 0.8796 & 0.9471 & 0.9841 & 0.9947 \\
& Deep K   & 0.8585 & 0.9114 & 0.9656 & 0.9881 \\
& Deep K/V & 0.9021 & 0.9524 & 0.9881 & 0.9947 \\
\midrule
\multirow{4}{*}{\shortstack[l]{Raven\\OLMo}}
& Aligned  & 0.8981 & 0.9352 & 0.9683 & 0.9854 \\
& Deep V   & 0.9325 & 0.9577 & 0.9815 & 0.9907 \\
& Deep K   & 0.8849 & 0.9431 & 0.9788 & 0.9907 \\
& Deep K/V & 0.9312 & 0.9656 & 0.9868 & 0.9947 \\
\midrule
\multirow{4}{*}{\shortstack[l]{Raven\\TinyLlama}}
& Aligned  & 0.8836 & 0.9272 & 0.9643 & 0.9894 \\
& Deep V   & 0.9246 & 0.9669 & 0.9894 & 0.9934 \\
& Deep K   & 0.8862 & 0.9312 & 0.9616 & 0.9947 \\
& Deep K/V & 0.9167 & 0.9590 & 0.9868 & 0.9974 \\
\bottomrule
\end{tabular}
\end{table}

\begin{figure}[htbp]
\centering
\includegraphics[width=.58\linewidth]{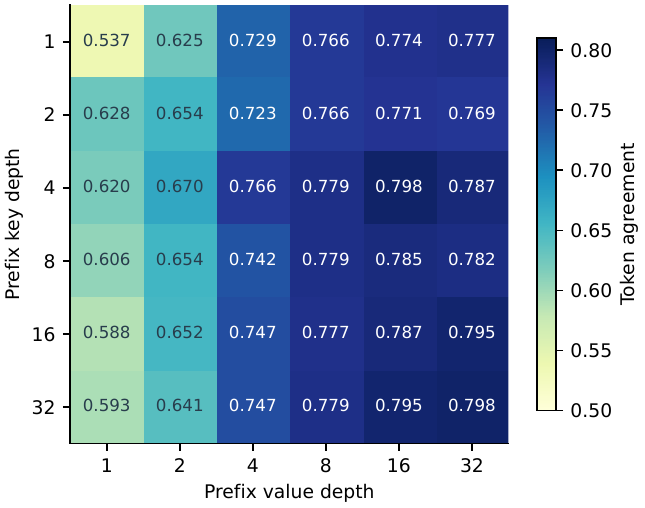}
\caption{
\textbf{Prefix read-depth grid.}
Huginn draft computation is fixed at $r=2$.
Rows vary verified-prefix K depth $d_K$ and columns vary V depth $d_V$;
cells report draft--target token agreement.
The four corners $d_K,d_V\in\{2,R\}$ correspond to Aligned, Deep V,
Deep K, and Deep K/V.
The diagnostic uses 8 prompts and 376 prediction positions, separate from
Table~\ref{tab:channels}.
}
\label{fig:grid}
\end{figure}

The separate Huginn depth grid in Figure~\ref{fig:grid} fixes the draft
computation at recurrent depth $r=2$ and varies the verified-prefix K and V
read depths independently.
The stronger variation along the V-depth axis supports the V-centric
motivation, while the remaining dependence on K depth motivates treating K
depth as a separate empirical choice.
This diagnostic uses a separate prompt cohort from Table~\ref{tab:channels}.

\subsection{Sequential Deep V and Deep K/V reads}

The motivation experiments isolate mature V as the strongest individual
channel intervention, whereas Mature-Spec uses a common Deep K/V policy in the
main system comparison.
For the three Raven checkpoints, Table~\ref{tab:kv-policy} compares
Deep V and Deep K/V configurations across all four tasks.
Joint Deep K/V gives the larger speedup and accepted length in every reported
setting, supporting the common Mature-Spec policy used in the main comparison.

\begin{table}[htbp]
\centering
\small
\caption{Sequential mature-prefix policies on the Raven checkpoints.
Entries report mean speedup [95\% CI] and accepted length $A$.}
\label{tab:kv-policy}
\setlength{\tabcolsep}{4pt}
\begin{tabular*}{\linewidth}{@{\extracolsep{\fill}}llcrcr@{}}
\toprule
Model & Task
& Deep V [95\% CI] & $A$
& Deep K/V [95\% CI] & $A$ \\
\midrule
Raven-Llama
& GSM8K
& 3.39$\times$ [3.25, 3.51] & 12.6
& 3.50$\times$ [3.37, 3.62] & 13.1 \\
& MATH-500
& 3.54$\times$ [3.36, 3.70] & 13.4
& 3.70$\times$ [3.55, 3.83] & 14.0 \\
& HumanEval
& 3.34$\times$ [3.12, 3.53] & 12.0
& 3.44$\times$ [3.22, 3.64] & 13.1 \\
& MBPP
& 3.72$\times$ [3.62, 3.81] & 13.3
& 3.84$\times$ [3.74, 3.92] & 14.7 \\
\midrule
Raven-OLMo
& GSM8K
& 3.50$\times$ [3.36, 3.64] & 13.0
& 3.63$\times$ [3.53, 3.73] & 13.5 \\
& MATH-500
& 3.55$\times$ [3.40, 3.69] & 13.1
& 3.73$\times$ [3.60, 3.86] & 13.8 \\
& HumanEval
& 3.11$\times$ [2.87, 3.34] & 10.9
& 3.26$\times$ [3.02, 3.48] & 12.1 \\
& MBPP
& 3.77$\times$ [3.67, 3.87] & 13.4
& 3.90$\times$ [3.82, 3.97] & 14.8 \\
\midrule
Raven-TinyLlama
& GSM8K
& 3.66$\times$ [3.51, 3.79] & 13.1
& 3.78$\times$ [3.67, 3.90] & 13.6 \\
& MATH-500
& 3.68$\times$ [3.55, 3.80] & 13.1
& 3.77$\times$ [3.64, 3.90] & 13.6 \\
& HumanEval
& 3.47$\times$ [3.24, 3.70] & 11.6
& 3.63$\times$ [3.39, 3.86] & 12.9 \\
& MBPP
& 3.94$\times$ [3.85, 4.02] & 13.3
& 4.04$\times$ [3.95, 4.12] & 14.5 \\
\bottomrule
\end{tabular*}
\end{table}

\subsection{Fixed-setting reads and latent carry}

Table~\ref{tab:component-ci} gives uncertainty estimates for the
fixed-setting component comparisons in Table~\ref{tab:components}.
The component and main-system studies use separate paired timing sessions;
all within-table comparisons retain their corresponding AR reference.

\begin{table}[h]
\centering
\footnotesize
\caption{
Fixed-setting component speedups with 95\% confidence intervals.
Point estimates correspond to the component-session measurements in
Table~\ref{tab:components}.
}
\label{tab:component-ci}
\setlength{\tabcolsep}{2.5pt}
\begin{tabular}{@{}lllcc@{}}
\toprule
Model & Procedure & Prefix
& GSM8K: $S$ [95\% CI]
& MATH-500: $S$ [95\% CI] \\
\midrule
Huginn
& Sequential & Aligned
& 1.54$\times$ [1.46, 1.63]
& 1.85$\times$ [1.69, 2.03] \\
& Sequential & Deep K/V
& 2.26$\times$ [2.16, 2.36]
& 2.42$\times$ [2.28, 2.56] \\
& Parallel, no carry & Aligned
& 0.65$\times$ [0.62, 0.68]
& 1.16$\times$ [0.93, 1.45] \\
& Parallel, no carry & Deep K/V
& 0.93$\times$ [0.89, 0.97]
& 1.06$\times$ [0.98, 1.15] \\
& Parallel + carry & Aligned
& 0.71$\times$ [0.68, 0.75]
& 1.20$\times$ [1.00, 1.45] \\
& Parallel + carry & Deep V
& 1.07$\times$ [1.02, 1.14]
& 1.66$\times$ [1.43, 1.94] \\
& Parallel + carry & Deep K/V
& 4.11$\times$ [3.90, 4.32]
& 4.63$\times$ [4.33, 4.94] \\
\midrule
Raven-Llama
& Sequential & Aligned
& 3.12$\times$ [2.96, 3.28]
& 3.18$\times$ [3.00, 3.35] \\
& Sequential & Deep K/V
& 3.50$\times$ [3.38, 3.63]
& 3.67$\times$ [3.52, 3.81] \\
& Parallel, no carry & Aligned
& 2.13$\times$ [1.94, 2.34]
& 2.97$\times$ [2.57, 3.39] \\
& Parallel, no carry & Deep K/V
& 2.64$\times$ [2.44, 2.87]
& 3.34$\times$ [2.93, 3.75] \\
& Parallel + carry & Aligned
& 2.66$\times$ [2.42, 2.92]
& 3.16$\times$ [2.79, 3.55] \\
& Parallel + carry & Deep V
& 3.52$\times$ [3.26, 3.78]
& 4.52$\times$ [4.16, 4.88] \\
& Parallel + carry & Deep K/V
& 5.40$\times$ [5.14, 5.66]
& 5.99$\times$ [5.70, 6.29] \\
\bottomrule
\end{tabular}
\end{table}

\subsection{Complete carry/K/V factorial}

To examine the interaction between latent carry and mature prefix reads, we
evaluate all eight combinations of carry, Deep K, and Deep V at the same
parallel drafting budget.
For this analysis, we additionally measure committed generation progress $G$.
For verification event $b$, let $u_b\in\{0,1\}$ denote whether a
target-provided correction or bonus token is retained after verification.
For prompt $p$ with $B_p>0$ verification events,
\begin{equation}
G_p
=
\frac{1}{B_p}\sum_b (a_b+u_b),
\qquad
G=\operatorname{mean}_p G_p .
\label{eq:committed-progress}
\end{equation}
Thus, $G$ is the mean number of committed tokens per verification, including
accepted draft tokens and target-provided correction or bonus tokens when
available.

Let $Y_{c,d_K,d_V}$ denote $G$ under binary indicators
$c,d_K,d_V\in\{0,1\}$ for carry, Deep K, and Deep V.
We measure the three-way factorial interaction
\begin{align}
E_{cKV}
=
\tfrac14\big[
&(Y_{111}-Y_{110}-Y_{101}+Y_{100})
\nonumber\\
&-(Y_{011}-Y_{010}-Y_{001}+Y_{000})
\big].
\label{eq:interaction}
\end{align}
The first term measures the K/V interaction with carry enabled, and the
second measures the same interaction without carry.
A positive $E_{cKV}$ therefore indicates that carry strengthens the
complementarity between mature K and V.

The interaction is positive across all evaluated model--task settings.
For Huginn and Raven-Llama, $E_{cKV}$ is $2.85$ and $1.20$ tokens on GSM8K,
and $2.76$ and $0.83$ tokens on MATH-500, respectively.
The Raven-OLMo and Raven-TinyLlama controls show the same qualitative pattern. Table~\ref{tab:factorial} reports the complete carry/K/V factorial results
from which these interaction effects are computed.

\begin{table}[h]
\centering
\footnotesize
\caption{
Complete carry/K/V factorial.
Entries report mean committed generation progress $G$ at the fixed parallel
drafting budget.
}
\label{tab:factorial}
\setlength{\tabcolsep}{3pt}
\begin{tabular*}{\linewidth}{@{\extracolsep{\fill}}ccc*{4}{rr}@{}}
\toprule
& & &
\multicolumn{2}{c}{Huginn} &
\multicolumn{2}{c}{Raven-Llama} &
\multicolumn{2}{c}{Raven-OLMo} &
\multicolumn{2}{c}{Raven-TinyLlama} \\
\cmidrule(lr){4-5}
\cmidrule(lr){6-7}
\cmidrule(lr){8-9}
\cmidrule(l){10-11}
Carry & K & V
& GSM & MATH
& GSM & MATH
& GSM & MATH
& GSM & MATH \\
\midrule
0 & 0 & 0 & 2.38 & 4.37 & 6.78 & 9.59 & 7.73 & 8.33 & 9.07 & 8.46 \\
0 & 0 & 1 & 3.14 & 4.81 & 7.98 & 10.46 & 9.57 & 10.08 & 10.46 & 9.31 \\
0 & 1 & 0 & 2.11 & 2.69 & 6.63 & 9.04 & 8.04 & 7.94 & 8.72 & 8.44 \\
0 & 1 & 1 & 3.42 & 3.95 & 8.50 & 10.84 & 9.50 & 10.53 & 10.31 & 9.65 \\
1 & 0 & 0 & 2.61 & 4.51 & 8.60 & 10.28 & 9.52 & 9.43 & 10.55 & 10.46 \\
1 & 0 & 1 & 3.99 & 6.25 & 11.45 & 14.84 & 13.19 & 14.66 & 12.96 & 14.16 \\
1 & 1 & 0 & 3.26 & 4.71 & 9.60 & 11.06 & 10.39 & 9.41 & 11.35 & 10.76 \\
1 & 1 & 1 & 16.60 & 18.30 & 17.97 & 19.88 & 18.26 & 19.13 & 19.08 & 19.03 \\
\bottomrule
\end{tabular*}
\end{table}

\subsection{Retuned carried controls}

Aligned and Deep V carried policies are also tuned independently over their
development grids.
Table~\ref{tab:retuned} reports the selected GSM8K configurations and
confirmation measurements.
This complements the fixed-budget intervention by checking whether the weaker
read policies can recover their gap through a different proposal budget.
Retuning improves both alternatives, but their best measured configurations
remain below the full Deep K/V DAS-Wave decoder on both models.

\begin{table}[h]
\centering
\small
\caption{Independently retuned carried controls on GSM8K.}
\label{tab:retuned}
\setlength{\tabcolsep}{5pt}
\begin{tabular}{@{}llrrrrr@{}}
\toprule
Model & Prefix policy & $\gamma$ & $J$ & $k$ & Speedup & $A$ \\
\midrule
\multirow{2}{*}{Huginn}
& Aligned & 8 & 4 & 2 & 1.49$\times$ & 1.29 \\
& Deep V & 32 & 6 & 2 & 1.84$\times$ & 2.28 \\
\midrule
\multirow{2}{*}{\shortstack[l]{Raven\\Llama}}
& Aligned & 16 & 12 & 1 & 2.94$\times$ & 5.40 \\
& Deep V & 16 & 12 & 2 & 3.70$\times$ & 8.64 \\
\bottomrule
\end{tabular}
\end{table}

\section{Refinement-budget sensitivity}
\label{app:budgets}

\subsection{Recurrent updates and verifier capacity}

Table~\ref{tab:sensitivity} contains the numerical values underlying
Figure~\ref{fig:sensitivity}.
These diagnostics use 20 GSM8K prompts and hold the remaining Wave schedule
fixed.
Increasing $k$ adds recurrent computation to every active position, whereas
increasing $\gamma$ permits a wider verifier and, until the schedule limit, a
wider active window.

\begin{table}[htbp]
\centering
\small
\caption{One-dimensional refinement-budget sweeps.
Entries report mean speedup [95\% CI] and accepted length $A$.}
\label{tab:sensitivity}
\setlength{\tabcolsep}{6pt}
\begin{tabular}{@{}lrrcr@{}}
\toprule
Model & $k$ & $\gamma$ & Speedup [95\% CI] & $A$ \\
\midrule
\multirow{6}{*}{Huginn}
& 1 & 32 & 3.95$\times$ [3.68, 4.24] & 11.40 \\
& 2 & 32 & 4.33$\times$ [4.11, 4.55] & 16.49 \\
& 4 & 32 & 3.54$\times$ [3.34, 3.73] & 18.89 \\
& 8 & 32 & 2.37$\times$ [2.23, 2.50] & 19.56 \\
& 2 & 8  & 1.98$\times$ [1.93, 2.02] & 7.36 \\
& 2 & 16 & 3.58$\times$ [3.43, 3.72] & 13.67 \\
\midrule
\multirow{6}{*}{\shortstack[l]{Raven\\Llama}}
& 1 & 32 & 5.47$\times$ [5.11, 5.82] & 17.21 \\
& 2 & 32 & 4.67$\times$ [4.38, 4.96] & 18.46 \\
& 4 & 32 & 3.64$\times$ [3.46, 3.80] & 20.05 \\
& 8 & 32 & 2.33$\times$ [2.21, 2.45] & 19.55 \\
& 1 & 8  & 2.36$\times$ [2.31, 2.40] & 7.40 \\
& 1 & 16 & 4.33$\times$ [4.18, 4.48] & 13.96 \\
\bottomrule
\end{tabular}
\end{table}

\subsection{Initial active width}
\label{app:winit}

The initial-width diagnostic varies
$W_{\mathrm{init}}\in\{1,2,4,8\}$ while retaining the remaining schedule.
Changing the initial width also changes total active-position work, so this is
a schedule-sensitivity study rather than an equal-compute comparison.
Table~\ref{tab:winit} shows only modest variation across the tested widths,
with no consistent advantage from increasing the initial window.
We therefore retain $W_{\mathrm{init}}=2$ as a simple fixed default rather than
as a tuned optimum.

\begin{table}[htbp]
\centering
\footnotesize
\caption{Initial-width sensitivity.
Direct paired ratios are relative to the default $W_{\mathrm{init}}=2$.}
\label{tab:winit}
\setlength{\tabcolsep}{3pt}
\begin{tabular}{@{}lrccr@{}}
\toprule
Model & $W_{\mathrm{init}}$
& Speedup [95\% CI]
& Ratio to $W_{\mathrm{init}}{=}2$ [95\% CI]
& $A$ \\
\midrule
\multirow{4}{*}{Huginn}
& 1 & 4.30$\times$ [4.08, 4.55]
& 0.985 [0.964, 1.002] & 17.95 \\
& 2 & 4.38$\times$ [4.13, 4.65]
& 1.000 [1.000, 1.000] & 18.34 \\
& 4 & 4.39$\times$ [4.15, 4.66]
& 1.004 [0.981, 1.026] & 18.59 \\
& 8 & 4.36$\times$ [4.09, 4.66]
& 0.996 [0.964, 1.030] & 18.78 \\
\midrule
\multirow{4}{*}{\shortstack[l]{Raven\\Llama}}
& 1 & 5.61$\times$ [5.19, 6.03]
& 1.029 [0.975, 1.085] & 19.88 \\
& 2 & 5.49$\times$ [5.09, 5.90]
& 1.000 [1.000, 1.000] & 18.90 \\
& 4 & 5.58$\times$ [5.17, 5.99]
& 1.020 [0.979, 1.065] & 19.78 \\
& 8 & 5.62$\times$ [5.26, 5.97]
& 1.029 [1.002, 1.062] & 19.64 \\
\bottomrule
\end{tabular}
\end{table}

\FloatBarrier
\section{Performance uncertainty and proposal-cost diagnostics}
\label{app:checks}

\subsection{Main-result uncertainty}

Table~\ref{tab:system-ci} reports 95\% confidence intervals for the point
estimates in Table~\ref{tab:system}.
These intervals quantify uncertainty in each method's AR-relative speedup and
are not direct pairwise tests between accelerated methods.

\begin{table}[htbp]
\centering
\fontsize{7.8}{9.8}\selectfont
\caption{
Main-result uncertainty.
Entries are mean throughput speedup $S$ [95\% CI] over paired full-depth AR.
Token Recycling uses moving-block bootstrap over its fixed request stream;
all other intervals use paired prompt bootstrap.
}
\label{tab:system-ci}
\setlength{\tabcolsep}{2pt}
\begin{tabular*}{\linewidth}{@{\extracolsep{\fill}}llcccc@{}}
\toprule
Model & Method & GSM8K & MATH-500 & HumanEval & MBPP \\
\midrule

\multirow{7}{*}{Huginn}
& Lookahead
& 1.83$\times$ [1.76, 1.90]
& 1.84$\times$ [1.71, 1.98]
& 1.50$\times$ [1.35, 1.68]
& 1.40$\times$ [1.31, 1.51] \\

& PLD
& 1.54$\times$ [1.45, 1.66]
& 1.93$\times$ [1.66, 2.24]
& 1.61$\times$ [1.29, 1.99]
& 1.35$\times$ [1.18, 1.55] \\

& Token Recycling
& 2.81$\times$ [2.72, 2.92]
& 2.81$\times$ [2.69, 2.88]
& 2.74$\times$ [2.43, 3.08]
& 2.56$\times$ [2.39, 2.76] \\

& SAM-Decoding
& 1.63$\times$ [1.51, 1.77]
& 2.50$\times$ [1.95, 3.20]
& 2.01$\times$ [1.43, 2.79]
& 1.49$\times$ [1.30, 1.72] \\

& Self-speculative
& 1.70$\times$ [1.63, 1.76]
& 1.82$\times$ [1.73, 1.91]
& 1.64$\times$ [1.53, 1.76]
& 1.79$\times$ [1.71, 1.88] \\

& Mature-Spec
& 2.26$\times$ [2.16, 2.36]
& 2.41$\times$ [2.27, 2.55]
& 2.35$\times$ [2.17, 2.52]
& 2.31$\times$ [2.18, 2.43] \\

& \daswave{}
& 4.12$\times$ [3.91, 4.33]
& 4.64$\times$ [4.33, 4.96]
& 4.00$\times$ [3.62, 4.40]
& 4.20$\times$ [3.88, 4.51] \\
\midrule

\multirow{7}{*}{\shortstack[l]{Raven\\Llama}}
& Lookahead
& 1.69$\times$ [1.63, 1.74]
& 1.56$\times$ [1.46, 1.68]
& 1.40$\times$ [1.24, 1.58]
& 1.46$\times$ [1.37, 1.54] \\

& PLD
& 1.34$\times$ [1.29, 1.40]
& 2.14$\times$ [1.62, 2.73]
& 1.64$\times$ [1.34, 1.97]
& 2.10$\times$ [1.75, 2.49] \\

& Token Recycling
& 2.59$\times$ [2.52, 2.67]
& 2.51$\times$ [2.39, 2.60]
& 2.42$\times$ [2.20, 2.66]
& 2.49$\times$ [2.43, 2.57] \\

& SAM-Decoding
& 1.53$\times$ [1.47, 1.58]
& 2.34$\times$ [1.73, 3.08]
& 1.73$\times$ [1.41, 2.10]
& 2.43$\times$ [1.99, 2.96] \\

& Self-speculative
& 3.12$\times$ [2.96, 3.28]
& 3.19$\times$ [3.01, 3.37]
& 3.07$\times$ [2.84, 3.29]
& 3.53$\times$ [3.41, 3.64] \\

& Mature-Spec
& 3.50$\times$ [3.37, 3.62]
& 3.70$\times$ [3.55, 3.83]
& 3.44$\times$ [3.22, 3.64]
& 3.84$\times$ [3.74, 3.92] \\

& \daswave{}
& 5.36$\times$ [5.09, 5.62]
& 6.01$\times$ [5.70, 6.31]
& 5.21$\times$ [4.74, 5.68]
& 6.28$\times$ [6.07, 6.49] \\
\midrule

\multirow{7}{*}{\shortstack[l]{Raven\\OLMo}}
& Lookahead
& 1.87$\times$ [1.79, 1.94]
& 1.70$\times$ [1.58, 1.84]
& 1.33$\times$ [1.17, 1.52]
& 1.84$\times$ [1.67, 2.03] \\

& PLD
& 1.44$\times$ [1.36, 1.52]
& 2.19$\times$ [1.70, 2.79]
& 1.58$\times$ [1.25, 2.02]
& 2.09$\times$ [1.77, 2.49] \\

& Token Recycling
& 2.80$\times$ [2.70, 2.92]
& 2.79$\times$ [2.66, 2.92]
& 2.40$\times$ [2.11, 2.65]
& 3.19$\times$ [2.90, 3.42] \\

& SAM-Decoding
& 1.65$\times$ [1.55, 1.75]
& 1.72$\times$ [1.52, 1.99]
& 2.12$\times$ [1.41, 3.19]
& 1.94$\times$ [1.61, 2.32] \\

& Self-speculative
& 3.31$\times$ [3.19, 3.43]
& 3.30$\times$ [3.13, 3.46]
& 2.82$\times$ [2.54, 3.10]
& 3.60$\times$ [3.46, 3.72] \\

& Mature-Spec
& 3.63$\times$ [3.53, 3.73]
& 3.73$\times$ [3.60, 3.86]
& 3.26$\times$ [3.02, 3.48]
& 3.90$\times$ [3.82, 3.97] \\

& \daswave{}
& 5.54$\times$ [5.34, 5.74]
& 5.96$\times$ [5.61, 6.31]
& 4.82$\times$ [4.25, 5.40]
& 6.34$\times$ [6.03, 6.64] \\
\midrule

\multirow{7}{*}{\shortstack[l]{Raven\\TinyLlama}}
& Lookahead
& 1.92$\times$ [1.86, 1.99]
& 1.47$\times$ [1.41, 1.55]
& 1.52$\times$ [1.37, 1.68]
& 1.62$\times$ [1.53, 1.71] \\

& PLD
& 1.95$\times$ [1.71, 2.27]
& 1.47$\times$ [1.36, 1.59]
& 2.64$\times$ [1.97, 3.37]
& 2.18$\times$ [1.92, 2.46] \\

& Token Recycling
& 2.78$\times$ [2.71, 2.86]
& 2.48$\times$ [2.36, 2.54]
& 2.45$\times$ [2.13, 2.66]
& 2.56$\times$ [2.45, 2.63] \\

& SAM-Decoding
& 1.88$\times$ [1.70, 2.13]
& 1.75$\times$ [1.49, 2.08]
& 2.78$\times$ [2.08, 3.60]
& 2.45$\times$ [2.12, 2.82] \\

& Self-speculative
& 3.44$\times$ [3.30, 3.58]
& 3.42$\times$ [3.25, 3.58]
& 3.26$\times$ [2.98, 3.53]
& 3.72$\times$ [3.58, 3.84] \\

& Mature-Spec
& 3.78$\times$ [3.67, 3.90]
& 3.77$\times$ [3.64, 3.90]
& 3.63$\times$ [3.39, 3.86]
& 4.04$\times$ [3.95, 4.12] \\

& \daswave{}
& 6.34$\times$ [6.08, 6.60]
& 6.39$\times$ [6.13, 6.65]
& 6.02$\times$ [5.49, 6.52]
& 6.96$\times$ [6.70, 7.20] \\
\bottomrule
\end{tabular*}
\end{table}

\subsection{Sequential proposal cost}

Table~\ref{tab:cost-example} provides a cost diagnostic for the selected
Huginn sequential drafters.
At similar accepted length, the depth-4 Mature-Spec draft substantially
reduces proposal time relative to depth-8 self-speculation, illustrating how
mature prefix reads can trade representation quality for shallower draft
computation.
This diagnostic is separate from the main timing sessions.

\begin{table}[htbp]
\centering
\small
\caption{
Huginn/GSM8K sequential proposal-cost diagnostic.
Draft time is the cumulative CUDA-event time spent on proposal construction
per request, excluding prefill and target verification.
}
\label{tab:cost-example}
\setlength{\tabcolsep}{6pt}
\begin{tabular}{@{}lrrrr@{}}
\toprule
Draft & $d$ & $A$ & Draft ms/request & ms/token \\
\midrule
Self-speculative & 8 & 5.2 & 4070 & 50.7 \\
Mature-Spec & 4 & 5.1 & 2293 & 36.1 \\
\bottomrule
\end{tabular}
\end{table}


\begin{thebibliography}{25}
\providecommand{\natexlab}[1]{#1}
\providecommand{\url}[1]{\texttt{#1}}
\expandafter\ifx\csname urlstyle\endcsname\relax
  \providecommand{\doi}[1]{doi: #1}\else
  \providecommand{\doi}{doi: \begingroup \urlstyle{rm}\Url}\fi

\bibitem[Austin et~al.(2021)Austin, Odena, Nye, Bosma, Michalewski, Dohan, Jiang, Cai, Terry, Le, et~al.]{austin2021}
Jacob Austin, Augustus Odena, Maxwell Nye, Maarten Bosma, Henryk Michalewski, David Dohan, Ellen Jiang, Carrie Cai, Michael Terry, Quoc Le, et~al.
\newblock Program synthesis with large language models.
\newblock \emph{arXiv preprint arXiv:2108.07732}, 2021.

\bibitem[Bae et~al.(2025)Bae, Fisch, Harutyunyan, Ji, Kim, and Schuster]{bae2025relaxed}
Sangmin Bae, Adam Fisch, Hrayr Harutyunyan, Ziwei Ji, Seungyeon Kim, and Tal Schuster.
\newblock Relaxed recursive transformers: Effective parameter sharing with layer-wise lora.
\newblock In \emph{International Conference on Learning Representations}, volume 2025, pp.\  34282--34327, 2025.

\bibitem[Chen et~al.(2023)Chen, Borgeaud, Irving, Lespiau, Sifre, and Jumper]{chen2023spec}
Charlie Chen, Sebastian Borgeaud, Geoffrey Irving, Jean-Baptiste Lespiau, Laurent Sifre, and John Jumper.
\newblock Accelerating large language model decoding with speculative sampling.
\newblock \emph{arXiv preprint arXiv:2302.01318}, 2023.

\bibitem[Chen et~al.(2021)Chen, Tworek, Jun, Yuan, Pinto, Kaplan, Edwards, Burda, Joseph, Brockman, et~al.]{chen2021code}
Mark Chen, Jerry Tworek, Heewoo Jun, Qiming Yuan, Henrique Ponde De~Oliveira Pinto, Jared Kaplan, Harri Edwards, Yuri Burda, Nicholas Joseph, Greg Brockman, et~al.
\newblock Evaluating large language models trained on code.
\newblock \emph{arXiv preprint arXiv:2107.03374}, 2021.

\bibitem[Cho et~al.(2026)Cho, Cui, Kim, Han, and Han]{loopspec2026}
SangLyul Cho, Langqing Cui, Sehoon Kim, Dongsu Han, and Insu Han.
\newblock Loopspec: Pipelined self-speculative decoding for looped transformers.
\newblock \emph{arXiv preprint arXiv:2609.17184}, 2026.

\bibitem[Cobbe et~al.(2021)Cobbe, Kosaraju, Bavarian, Chen, Jun, Kaiser, Plappert, Tworek, Hilton, Nakano, et~al.]{cobbe2021}
Karl Cobbe, Vineet Kosaraju, Mohammad Bavarian, Mark Chen, Heewoo Jun, Lukasz Kaiser, Matthias Plappert, Jerry Tworek, Jacob Hilton, Reiichiro Nakano, et~al.
\newblock Training verifiers to solve math word problems.
\newblock \emph{arXiv preprint arXiv:2110.14168}, 2021.

\bibitem[Dehghani et~al.(2018)Dehghani, Gouws, Vinyals, Uszkoreit, and Kaiser]{dehghani2019universal}
Mostafa Dehghani, Stephan Gouws, Oriol Vinyals, Jakob Uszkoreit, and {\L}ukasz Kaiser.
\newblock Universal transformers.
\newblock \emph{arXiv preprint arXiv:1807.03819}, 2018.

\bibitem[Elhoushi et~al.(2024)Elhoushi, Shrivastava, Liskovich, Hosmer, Wasti, Lai, Mahmoud, Acun, Agarwal, Roman, et~al.]{layerskip2024}
Mostafa Elhoushi, Akshat Shrivastava, Diana Liskovich, Basil Hosmer, Bram Wasti, Liangzhen Lai, Anas Mahmoud, Bilge Acun, Saurabh Agarwal, Ahmed Roman, et~al.
\newblock Layerskip: Enabling early exit inference and self-speculative decoding.
\newblock In \emph{Proceedings of the 62nd Annual Meeting of the Association for Computational Linguistics (Volume 1: Long Papers)}, pp.\  12622--12642, 2024.

\bibitem[Fu et~al.(2024)Fu, Bailis, Stoica, and Zhang]{lookahead2024}
Yichao Fu, Peter Bailis, Ion Stoica, and Hao Zhang.
\newblock Break the sequential dependency of llm inference using lookahead decoding.
\newblock \emph{arXiv preprint arXiv:2402.02057}, 2024.

\bibitem[Geiping et~al.(2025)Geiping, Yang, and Su]{psampler2025}
Jonas Geiping, Xinyu Yang, and Guinan Su.
\newblock Efficient parallel samplers for recurrent-depth models and their connection to diffusion language models.
\newblock \emph{arXiv preprint arXiv:2510.14961}, 2025.

\bibitem[Geiping et~al.(2026)Geiping, McLeish, Jain, Kirchenbauer, Singh, Bartoldson, Kailkhura, Bhatele, and Goldstein]{geiping2025scaling}
Jonas Geiping, Sean McLeish, Neel Jain, John Kirchenbauer, Siddharth Singh, Brian Bartoldson, Bhavya Kailkhura, Abhinav Bhatele, and Tom Goldstein.
\newblock Scaling up test-time compute with latent reasoning: A recurrent depth approach.
\newblock \emph{Advances in Neural Information Processing Systems}, 38:\penalty0 41340--41391, 2026.

\bibitem[Giannou et~al.(2023)Giannou, Rajput, Sohn, Lee, Lee, and Papailiopoulos]{giannou2023looped}
Angeliki Giannou, Shashank Rajput, Jy-yong Sohn, Kangwook Lee, Jason~D Lee, and Dimitris Papailiopoulos.
\newblock Looped transformers as programmable computers.
\newblock In \emph{International Conference on Machine Learning}, pp.\  11398--11442. PMLR, 2023.

\bibitem[Hendrycks et~al.(2021)Hendrycks, Burns, Kadavath, Arora, Basart, Tang, Song, and Steinhardt]{hendrycks2021}
Dan Hendrycks, Collin Burns, Saurav Kadavath, Akul Arora, Steven Basart, Eric Tang, Dawn Song, and Jacob Steinhardt.
\newblock Measuring mathematical problem solving with the math dataset.
\newblock \emph{arXiv preprint arXiv:2103.03874}, 2021.

\bibitem[Hooper et~al.(2025)Hooper, Kim, Mohammadzadeh, Genc, Keutzer, Gholami, and Sophia~Shao]{speed2023}
Coleman Hooper, Sehoon Kim, Hiva Mohammadzadeh, Hasan Genc, Kurt Keutzer, Amir Gholami, and Yakun Sophia~Shao.
\newblock Speed: Speculative pipelined execution for efficient decoding.
\newblock In \emph{Enhancing LLM Performance: Efficacy, Fine-Tuning, and Inference Techniques}, pp.\  19--32. Springer, 2025.

\bibitem[Hu et~al.(2025)Hu, Wang, Zhang, Zhang, Li, Chen, and Zhang]{samd2025}
Yuxuan Hu, Ke~Wang, Xiaokang Zhang, Fanjin Zhang, Cuiping Li, Hong Chen, and Jing Zhang.
\newblock Sam decoding: Speculative decoding via suffix automaton.
\newblock In \emph{Proceedings of the 63rd Annual Meeting of the Association for Computational Linguistics (Volume 1: Long Papers)}, pp.\  12187--12204, 2025.

\bibitem[Leviathan et~al.(2023)Leviathan, Kalman, and Matias]{leviathan2023}
Yaniv Leviathan, Matan Kalman, and Yossi Matias.
\newblock Fast inference from transformers via speculative decoding.
\newblock In \emph{International conference on machine learning}, pp.\  19274--19286. PMLR, 2023.

\bibitem[Lewkowycz et~al.(2022)Lewkowycz, Andreassen, Dohan, Dyer, Michalewski, Ramasesh, Slone, Anil, Schlag, Gutman-Solo, et~al.]{lewkowycz2022}
Aitor Lewkowycz, Anders Andreassen, David Dohan, Ethan Dyer, Henryk Michalewski, Vinay Ramasesh, Ambrose Slone, Cem Anil, Imanol Schlag, Theo Gutman-Solo, et~al.
\newblock Solving quantitative reasoning problems with language models.
\newblock \emph{Advances in neural information processing systems}, 35:\penalty0 3843--3857, 2022.

\bibitem[Li et~al.(2026{\natexlab{a}})Li, Jiang, Chen, Li, Yu, Dong, Ren, Tang, and Yuan]{li2025scout}
Guanghao Li, Wenhao Jiang, Mingfeng Chen, Yan Li, Hao Yu, Shuting Dong, Tao Ren, Ming Tang, and Chun Yuan.
\newblock Scout: Teaching pre-trained language models to enhance reasoning via flow chain-of-thought.
\newblock \emph{Advances in Neural Information Processing Systems}, 38:\penalty0 95340--95364, 2026{\natexlab{a}}.

\bibitem[Li et~al.(2026{\natexlab{b}})Li, Jiang, Shen, Tang, and Yuan]{li2026efficient}
Guanghao Li, Wenhao Jiang, Li~Shen, Ming Tang, and Chun Yuan.
\newblock Efficient transformer parameter reuse via zero-token mechanism.
\newblock In \emph{Findings of the Association for Computational Linguistics: ACL 2026}, pp.\  14498--14515, 2026{\natexlab{b}}.

\bibitem[Lightman et~al.(2024)Lightman, Kosaraju, Burda, Edwards, Baker, Lee, Leike, Schulman, Sutskever, and Cobbe]{lightman2023}
Hunter Lightman, Vineet Kosaraju, Yuri Burda, Harrison Edwards, Bowen Baker, Teddy Lee, Jan Leike, John Schulman, Ilya Sutskever, and Karl Cobbe.
\newblock Let's verify step by step.
\newblock In \emph{International Conference on Learning Representations}, volume 2024, pp.\  39578--39601, 2024.

\bibitem[Luo et~al.(2025)Luo, Wang, Zhu, Zhang, Zhang, Yang, and Xu]{tokenrecycling2024}
Xianzhen Luo, Yixuan Wang, Qingfu Zhu, Zhiming Zhang, Xuanyu Zhang, Qing Yang, and Dongliang Xu.
\newblock Turning trash into treasure: Accelerating inference of large language models with token recycling.
\newblock In \emph{Proceedings of the 63rd Annual Meeting of the Association for Computational Linguistics (Volume 1: Long Papers)}, pp.\  6816--6831, 2025.

\bibitem[McLeish et~al.(2025)McLeish, Li, Kirchenbauer, Kalra, Bartoldson, Kailkhura, Schwarzschild, Geiping, Goldstein, and Goldblum]{mcleish2025retrofit}
Sean McLeish, Ang Li, John Kirchenbauer, Dayal~Singh Kalra, Brian~R Bartoldson, Bhavya Kailkhura, Avi Schwarzschild, Jonas Geiping, Tom Goldstein, and Micah Goldblum.
\newblock Teaching pretrained language models to think deeper with retrofitted recurrence.
\newblock \emph{arXiv preprint arXiv:2511.07384}, 2025.

\bibitem[Saxena(2023)]{pld2023}
Apoorv Saxena.
\newblock Prompt lookup decoding, November 2023.
\newblock URL \url{https://github.com/apoorvumang/prompt-lookup-decoding/}.

\bibitem[Schwethelm et~al.(2026)Schwethelm, Rueckert, and Kaissis]{cdb2026}
Kristian Schwethelm, Daniel Rueckert, and Georgios Kaissis.
\newblock Depth-adaptive inference of looped language models via continuous depth batching.
\newblock \emph{arXiv preprint arXiv:2608.09444}, 2026.

\bibitem[Zhang et~al.(2024)Zhang, Wang, Li, Shou, Chen, Chen, and Mehrotra]{draftverify2024}
Jun Zhang, Jue Wang, Huan Li, Lidan Shou, Ke~Chen, Gang Chen, and Sharad Mehrotra.
\newblock Draft\& verify: Lossless large language model acceleration via self-speculative decoding.
\newblock In \emph{Proceedings of the 62nd Annual Meeting of the Association for Computational Linguistics (Volume 1: Long Papers)}, pp.\  11263--11282, 2024.

\end{thebibliography}
\end{document}